\documentclass[english, 10pt, twocolumn, twoside]{IEEEtran}  % twocolumn

\usepackage{threeparttable}
\usepackage{cite}
\ifCLASSINFOpdf
\else
\fi

\usepackage{graphics}
\usepackage{epstopdf}
\usepackage[cmex10]{amsmath}
\usepackage{amsfonts}
\usepackage{cases}
\usepackage{booktabs}
\usepackage{epsfig}
\usepackage{url}
\usepackage{algorithm}
\usepackage{algorithmic}
\usepackage{array}
\usepackage{multirow}
\usepackage{bigstrut}
\usepackage{amssymb, bm}
\usepackage{color}
\usepackage{booktabs}
\usepackage{comment}
\usepackage{subcaption}

\def\etal{\emph{et al}.~}
\def\ie{\emph{i.e.}~}
\def\eg{\emph{e.g.}~}
\begin{document}

\title{STC-Flow: Spatio-temporal Context-aware Optical Flow Estimation}

\author{Xiaolin~Song,
        Yuyang Zhao,
        and Jingyu~Yang %,
        %Cuiling~Lan,
        %Wenjun~Zeng,
        %and~Jiahao~Li

        % <-this % stops a space

%\thanks{This work was supported by the National Natural Science Foundation of China (Grant No. 61771339), and by the Reserved Peiyang Scholar Program of Tianjin University, Tianjin, China. (\emph{Corresponding authors: Jingyu Yang.})}
%\thanks{X. Song and J. Yang are with the School of Electrical and Information Engineering, Tianjin University, Tianjin 300072, China (E-mail: \{songxl, yjy\}@tju.edu.cn).}
%\thanks{C. Lan and W. Zeng are with Microsoft Research Asia, Beijing 100080, China (E-mail: \{culan, wezeng\}@microsoft.com).}
%\thanks{J. Xing is with the National Laboratory of Pattern Recognition, Institute of Automation, Chinese Academy of Sciences, Beijing 100190, China (E-mail: jlxing@nlpr.ia.ac.cn).}
}% <-this

%\markboth{IEEE Transactions on Circuits and Systems for Video Technology}%
%{Song \MakeLowercase{\textit{et al.}}: STC-Flow: Spatio-temporal Context-aware Optical Flow Estimation}

\maketitle

\begin{abstract}
% 11 lines

In this paper, we propose a spatio-temporal contextual network, STC-Flow, for optical flow estimation. Unlike previous optical flow estimation approaches with local pyramid feature extraction and multi-level correlation, we propose a contextual relation exploration architecture by capturing rich long-range dependencies in spatial and temporal dimensions. Specifically, STC-Flow contains three key context modules --- pyramidal spatial context module, temporal context correlation module and recurrent residual contextual upsampling module, to build the relationship in each stage of feature extraction, correlation, and flow reconstruction, respectively. Experimental results indicate that the proposed scheme achieves the state-of-the-art performance of two-frame based methods on the Sintel dataset and the KITTI 2012/2015 datasets. %,  with improvement by 0.80, 1.15 and 0.10 in terms of average end-point error (AEE) against competing baseline methods --- FlowNet2, LiteFlowNet and PWC-Net on the \emph{Final} pass of Sintel dataset, respectively.
\end{abstract}

%\begin{IEEEkeywords}
%Pyramid correlation mapping, optical flow estimation, deep learning
%\end{IEEEkeywords}

\IEEEpeerreviewmaketitle

\section{Introduction}

Optical flow estimation is an important yet challenging problem in the field of video analytics. Recently, deep learning based approaches have been extensively exploited to estimate optical flow via convolutional neural networks (CNNs). Despite the great efforts and rapid developments, the advancements are not as significant as works in single image based computer vision tasks. The main reason is that optical flow is not directly measurable in the wild, and it is challenging to model motion dynamics with pixel-wise correspondence between two consecutive frames, which would contain variable motion displacements; thus optical flow estimation requires the efficient representation of features to match different motion objects or scenes.

\begin{figure}[t]
  \begin{center}
   \includegraphics[width=0.9\linewidth]{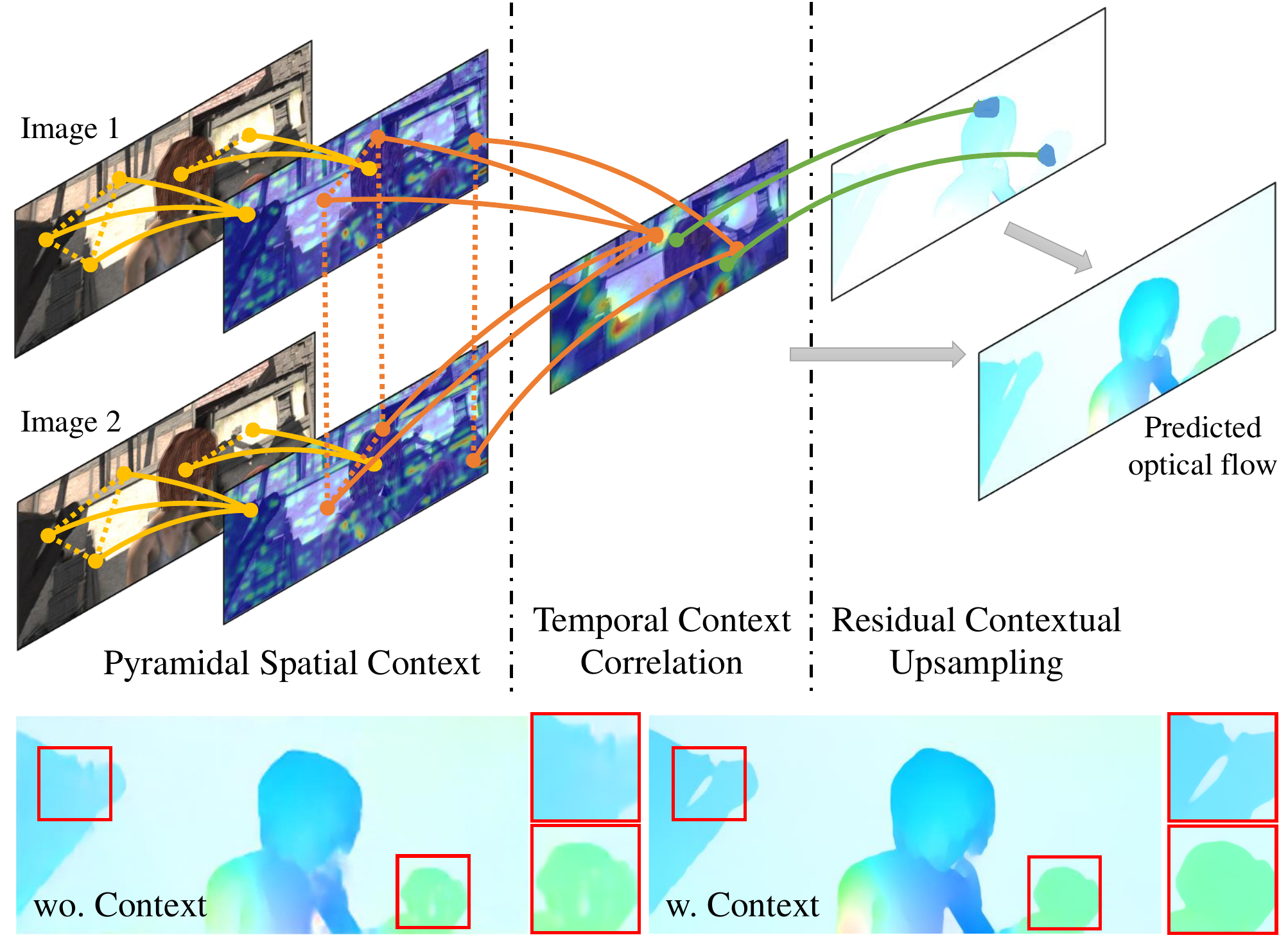}
  \end{center}
  % \vspace{-5mm}
  \caption{Overview of spatio-temporal contextual network for optical flow estimation. The context modules aim to build the relationship in spatial and temporal dimensions. With multiple context modeling, STC-Flow achieves the better performance with rich details.}
  \label{fig:intro}
  % \vspace{-6mm}
\end{figure}

Conventional methods attempt to propose mathematical algorithms of optical flow estimation such as DeepFlow \cite{Weinzaepfel2013DeepFlow} and EpicFlow \cite{revaud2015epicflow} by matching features of two frames. Most of these methods, however, are complicated with heavy computational complexity, and usually fail for motions with large displacements. CNN-based methods for optical flow estimation, which usually utilize encoder-decoder architectures with pyramidal feature extraction and flow reconstruction like FlowNet \cite{dosovitskiy2015flownet}, SpyNet \cite{ranjan2017optical}, PWC-Net \cite{Sun_2018_CVPR}, boost the state-of-the-art performance of optical flow estimation and outperform conventional methods. However, the features in lower level contain rich details, while the receptive field is small, which is not effective to catch the larger displacement of motions; while the features in higher level highlight the overall outlines or shapes of objects, with less details, which may cause the misalignment of objects with complex shapes or non-rigid motions. So it is essential to capture context information with large receptive field and long-range dependencies, to build the global relationship for each stage of CNNs.% the CNN-based network.

In this paper, as shown in Figure \ref{fig:intro}, we propose an end-to-end architecture for optical flow estimation, with jointly and effectively spatio-temporal contextual network. To respectively build the relationship in each stage of feature extraction, correlation and flow reconstruction, the network contains three key context modules: (a) \emph{Pyramidal spatial context module} aims to enhance the discriminant ability of feature representations in spatial dimension. (b) \emph{Temporal context correlation module} is adopted to model the global spatio-temporal relationships of the cost volume calculated from correlation operation. (c) \emph{Recurrent residual context upsampling module} leverages the underlying content of predicted flow field between adjacent levels, to learn high-frequency features and preserve edges within a large receptive field.

In summary, the main contributions of this work are three-fold:
%% \vspace{-2mm}
\begin{itemize}
\item We propose a general framework, \ie contextual attention framework, for efficient feature representation learning, which benefits multiple inputs and complicated target operation.% In addition, we propose an approach for simplified calculation of matrix multiplication for attention mechanism.%   pyramid correlation mapping operation for jointly embedding the cost volumes in different scales, to yield detailed motion information in multi-scale features for temporal modeling. And we present a CWN module for fusing features from backbone and reconstruction branch, to refine intrinsic flow estimation by optimizing warped feature and cost volume.
\item We propose corresponding context modules based on the contextual attention framework, for feature extraction, correlation and optical flow reconstruction stages. %a residual learning branch for optical flow reconstruction, to further reconstruct the sub-band high-frequency residuals of finer optical flow in each stage, which can better exploit the temporal dynamics to a more accurate and effective representation and occlusion compensation between two frames.
\item Our network achieves the state-of-the-art performance in the Sintel and KITTI datasets for two-frame based optical flow estimation. %We propose a novel semi-supervised method for optical flow estimation, which uses supervised as well as unsupervised learning cues.
\end{itemize}

%-------------------------------------------------------------------------
\section{Related Work}

\noindent\textbf{Optical flow estimation. }%Horn and Schunck \cite{horn1981determining} pioneer the study on optical flow estimation. Brox \etal \cite{brox2004high} take advantage of illumination changes by combining the brightness and propose the warping-based estimation method. Brox \etal \cite{brox2010large} aggregate rich descriptors with feature matching into the variational formulation. Weinzaepfel \etal \cite{Weinzaepfel2013DeepFlow} propose DeepFlow to correlate multi-scale patches to generate cost volume and incorporate correlation operation as the matching term in functional. Revaud \etal \cite{revaud2015epicflow} propose EpicFlow that uses externally matched flows as initialization and interpolates them to dense flow. Zimmer \etal \cite{zimmer2011optic} propose the anisotropic smoothness term for complementary regularization to the data term.
%With the development of deep learning, CNN-based methods achieve a breakthrough on optical flow estimation.
Inspired by the success of CNNs, various deep networks for optical flow estimation have been proposed.
Dosovitskiy \etal \cite{dosovitskiy2015flownet} establish FlowNet which is the important CNN exploration on optical flow estimation with the encoder-deconder architecture, of which FlowNetS and FlowNetC are proposed with simple operations. However, the number of parameters is large with heavy calculation on correlation. Ilg \etal \cite{Ilg_2017_CVPR} propose a cascaded network with milestone performance based on FlowNetS and FlowNetC with huge parameters and expensive computation complexity.%Some methods use CNN models for image patches matching. Thewlis \etal \cite{Thewlis2016Fully} utilize Deep Matching formulation as an end-to-end CNN. Gadot \cite{Gadot2016PatchBatch} and Bailer \etal \cite{bailer2017cnn} use patch matching for Siamese network architectures with heavy computing. Moreover, patch matching based methods lack the capacity to apply to the larger context of the entire image because of the small image patches operator. Ranjan \etal \cite{ranjan2017optical} present a compact network named SPyNet inspired from spatial pyramid with multi-level representation learning. Nevertheless, the performance is not significant. with high accuracy for optical flow estimation, They utilize light feature-level matching and warping motivated by conventional methods.

To reduce the number of parameters,Ranjan \etal \cite{ranjan2017optical} present a compact SPyNet for spatial pyramid with multi-level representation learning. Hui \etal \cite{hui2018liteflownet} propose LiteFlowNet and Sun \etal \cite{Sun_2018_CVPR} propose PWC-Net, which are pioneers of the trend to lightweight optical flow estimation networks. LiteFlowNet \cite{hui2018liteflownet} involves cascaded flow inference for flow warping and feature matching. PWC-Net \cite{Sun_2018_CVPR} utilizes feature pyramid extraction and feature warping to construct the cost volume, and uses context network for optical flow refinement. HD$^3$ \cite{yin2018hierarchical} decomposes the full match density into hierarchical features to estimate the local matching, with heavy computational complexity. IRR \cite{hur2019iterative} involves the iterative residual refinement scheme, and integrates occlusion prediction as additional auxiliary supervision. SelFlow \cite{liu2019selflow} uses reliable flow predictions from non-occluded pixels, to learn optical flow for hallucinated occlusions from multiple frames for better performance.  % Our method introduces an unsupervised term, a flow regularization term for semi-supervision and a global refinement branch to learn reconstruction residuals in each stage for accurate flow estimation inspired by conventional methods and CNN-based methods.all utilized with light feature-level matching and warping,, and feature-driven local convolution (f-lconv) for flow regularization

\begin{figure*}[t]
 \centering\includegraphics[width=0.9\textwidth]{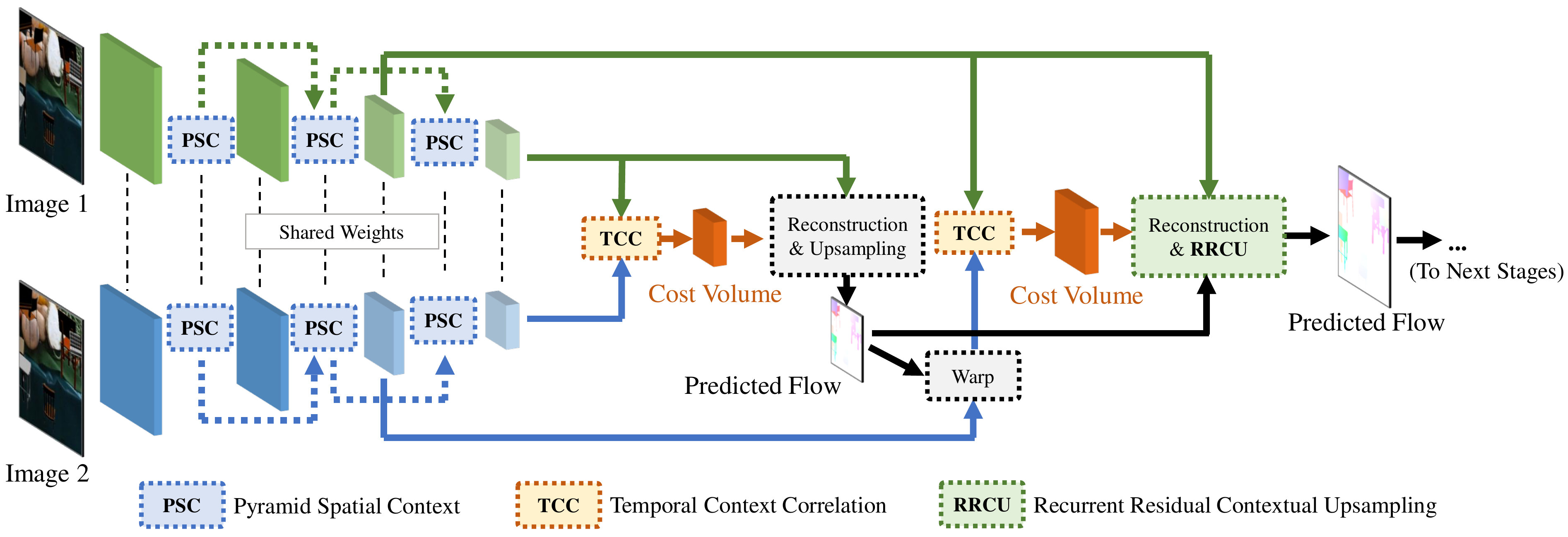}
 % \vspace{-2mm}
 \caption{The overall architecture of our spatio-temporal contextual network for optical flow estimation (STC-Flow). PSC, TCC and RRCU modules are flexible to adopt to model relationships of intra-/extra-features in each stage. These modules at only the top two levels are shown.} % ``PSC'' ``TCC'' and ``RRCU'' denote the three contextual modules --- pyramidal spatial context, temporal context correlation and recurrent residual contextual upsampling, respectively.  A 4-stage design with 1 stage of all contextual modules (PSC, TCC and RRCU) are shown.
 \label{fig:arch}
 % \vspace{-5mm}
\end{figure*}

\noindent\textbf{Context modeling in neural networks. } Context modeling has been successfully applied to capture long-range dependencies. Since a typical convolution operator has a local receptive field, context learning can affect an individual element by aggregating information from all elements. Many recent works utilize spatial self-attention to emphasize features of the key local regions \cite{du2018interaction-aware,zhang2019self-attention}. Object relation module \cite{hu2018relation} extends original attention to geometric relationship, and could be applied to improve the performance of object detection and other tasks. DANet \cite{fu2018dual} and CBAM \cite{woo2018cbam} introduce the channel-wise attention via self-attention mechanism. Global context network \cite{cao2019gcnet} effectively models the global context with a lightweight architecture. Specifically, the non-local network \cite{wang2017non-local} utilizes 3D convolution layers to aggregate spatial and temporal long-range dependencies for video frames.

In the optical flow estimation task, spatial contextual information helps to refine details and deal with occlusion. PWC-Net \cite{Sun_2018_CVPR} consists of the context network with stacked dilated convolution layers for flow post-processing. In LiteFlowNet \cite{hui2018liteflownet}, flow regularization layer is applied to ameliorate the issue of outliers and fake edges. IRR \cite{hur2019iterative} utilizes bilateral filters to refine blurry flow and occlusion. Nevertheless, previous work of context modeling in optical flow estimation mainly focuses on spatial features. For motion context modeling, it is essential to provide an elegant framework to explore spatial and temporal information. Accordingly, our network introduces spatial and temporal context module, and also introduce recurrent context to upsample spatial features of predicted flow field.
% Non-local, relation, Self-att, GCnet, DANet, CCNet ...

\section{STC-Flow}

Given a pair of video frames, scene or objects are diverse on movement velocity and direction in temporal dimension, and changes in scales, views, and luminance in spatial dimension. Convolutional operations built in CNNs process just a local neighborhood, and thus convolution stacks would lead to a local receptive field. The features corresponding to the pixels have similar textures of one object, even though they have differences in motion. These textures would introduce false-positive correlation, which result in wrong prediction of optical flow.

 To address this issue, our method, \ie STC-Flow, models contextual information by building global associations of intra-/extra-features with the attention mechanism in spatial and temporal dimensions respectively. The network could adaptively aggregate long-range contextual information, thus optimizing feature representation in feature extraction, correlation, and reconstruction stages, as shown in Figure \ref{fig:arch}.
In this section, we first introduce the contextual attention framework with single or multiple inputs for efficient feature representation learning. Based on the framework, we then propose three key context modules: \emph{pyramidal spatial context (PSC) module}, \emph{temporal context correlation  (TCC) module} and \emph{recurrent residual contextual upsampling (RRCU) module} for modeling contextual information.

%\subsection{Analysis on Attention Mechanism}
\subsection{Contextual Attention Framework}

\begin{figure*}[t]
 \centering\includegraphics[width=0.85\linewidth]{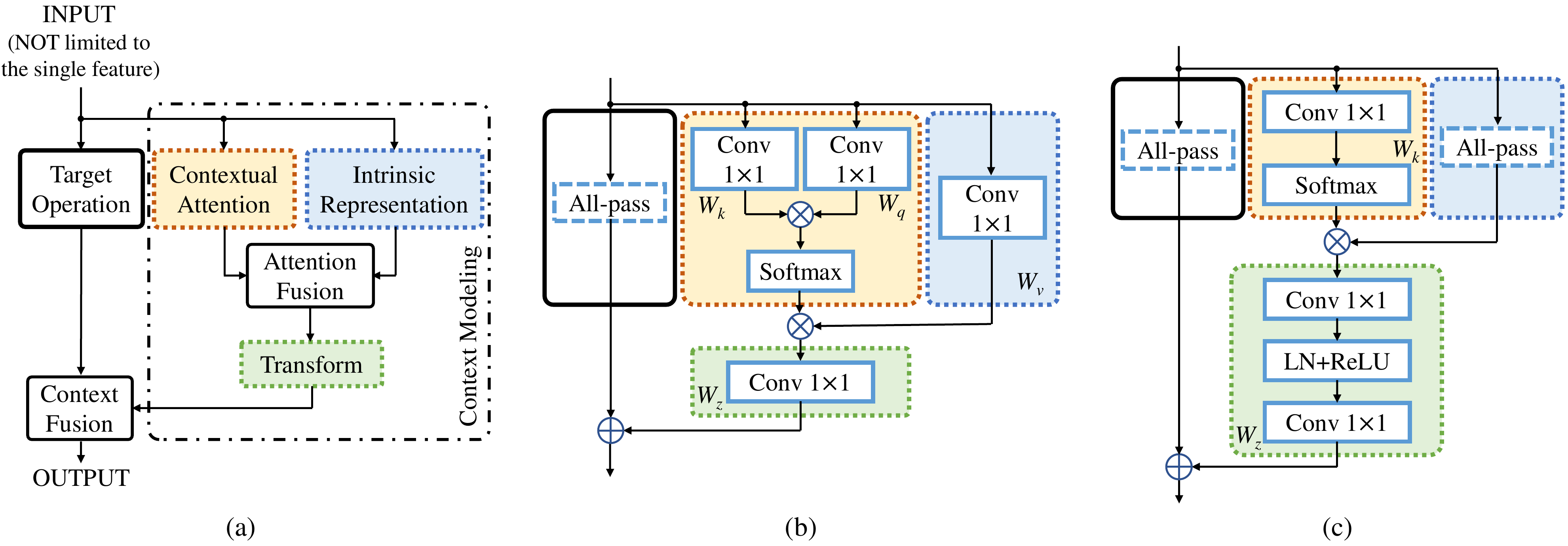}
 % \vspace{-2mm}
 \caption{The contextual attention framework (a) with modularization; and the specified forms of (b) the non-local block \cite{wang2017non-local}, and (c) global context (GC) block \cite{cao2019gcnet}.}
 \label{fig:caf}
 % \vspace{-3mm}
\end{figure*}

\noindent\textbf{Analysis on Attention Mechanism.} To capture long-range dependencies and model contextual details of single images or video clips, the basic non-local network \cite{wang2017non-local} aggregates pixel-wise information via self-attention mechanism. We denote $X$ and $Z$ as the input and output signals, such as the single image and the video clip. The non-local block can be expressed as follows:

\begin{footnotesize}
\begin{equation}
Z_i  = X_i  + W_z \sum\limits_j {\frac{{f(X_i ,X_j )}}{{\mathcal{N}(X)}}} (W_v X_j ),
\end{equation}
\end{footnotesize}

\noindent where $i$ and $j$ are the indices of target position coordinates and all possible enumerated positions. $f(X_i ,X_j )$ denotes the relationship between position $i$ and $j$, which is normalized by a factor $\mathcal{N}(X)$. The matrix multiplication operation is utilized to strengthen details of each query position. Embedded Gaussian is a widely-used instantiation of $f(X_i ,X_j )$, to compute similarity in an embedding space, and normalized with a softmax function, which is a soft selection across channels in one position. the non-local block with Embedded Gaussian is shown in Figure \ref{fig:caf}(b), and is expressed as follows: % to force the sum of values in one dimension to 1

\begin{footnotesize}
\begin{equation}
Z_i  = X_i  + W_z \sum\limits_j {\frac{{\exp((W_q X_i)^{\top}(W_k X_j ))}}{{\sum\limits_m {\exp((W_q X_i)^{\top}(W_k X_m ))}}}} (W_v X_j ),
\end{equation}
\end{footnotesize}

\noindent where $W_q$, $W_k$ and $W_v$ are linear transformation matrices.
%\textbf{Discussions.} Both two type of matrix multiplication.

\begin{figure}[t]
 \centering\includegraphics[width=0.7\linewidth]{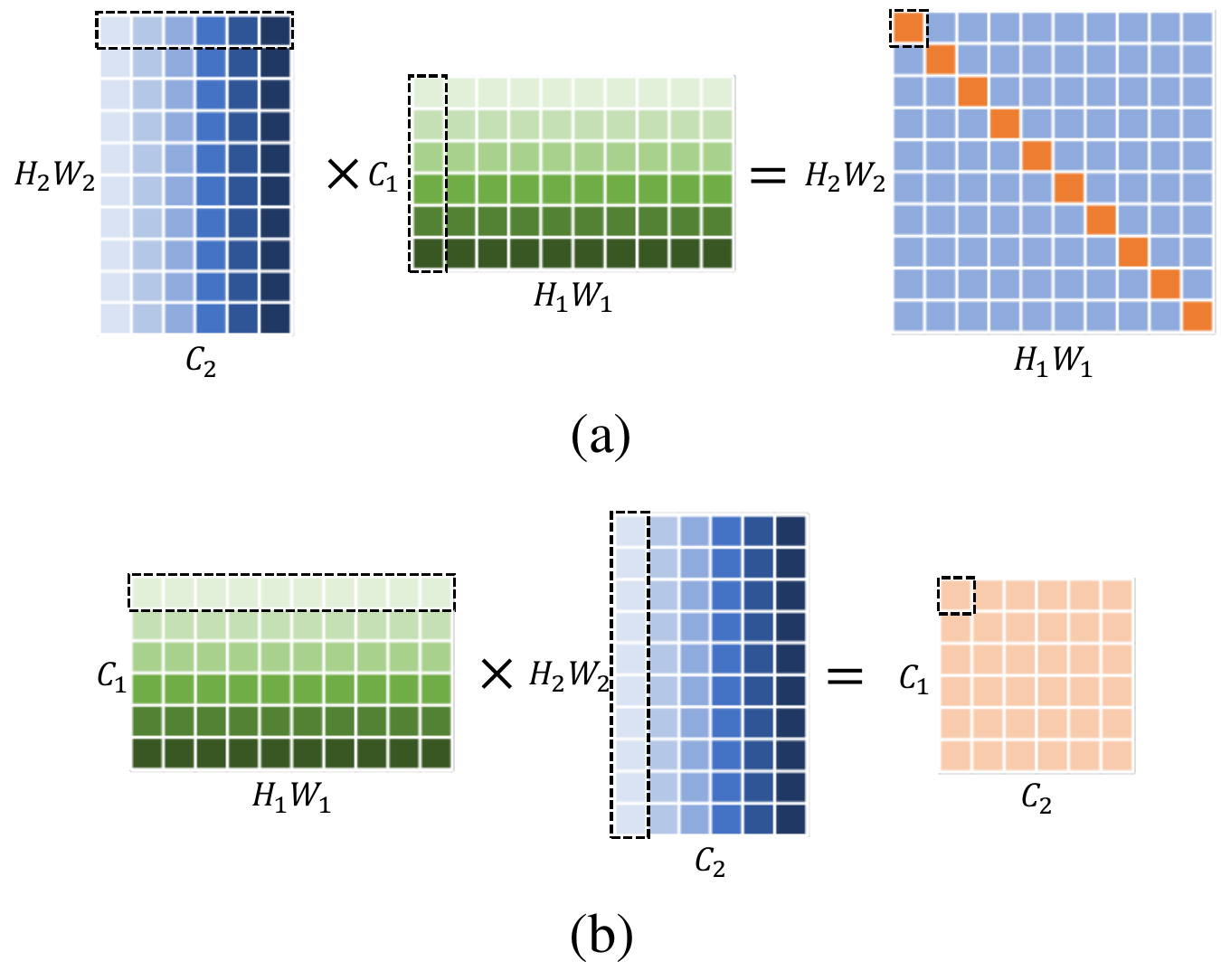}
 % \vspace{-2mm}
 \caption{The matrix multiplication with different contextual information. (a) The position-wise attention embedding; (b) the channel-wise embedding, also the global correlation of feature pairs.}
 \label{fig:twotypesmulti}
 % \vspace{-3mm}
\end{figure}

\noindent\textbf{Why attention for optical flow estimation?} Here, we discuss the relation between correlation in optical flow estimation and matrix multiplication in self-attention mechanism. We aim to explore the contextual information from the input pairs. Denote the feature pairs by $F_1$ and $F_2$. To distinguish the features from different input paths, we mark the size of features with height $H_k$, width $W_k$, channel $C_k$ ($k=1,2$), and position coordinate $\mathbf{x}_k \in [1,H_k] \times [1,W_k]$, channel index $c_k \in C_k$, and here $H_1=H_2$, $W_1=W_2$ and $C_1=C_2$. As the key function in optical flow estimation, the ``correlation'' operation between two patches of feature pairs, $\mathbf{f}_1$ and $\mathbf{f}_2$, is defined as follows for temporal modeling:

\begin{footnotesize}
\begin{equation}
\begin{split}
Corr (\mathbf{f}_{1},\mathbf{f}_{2})= \sum\limits_{\mathbf{o}} {\left\langle\mathbf{f}_{1} (\mathbf{x}_1+\mathbf{o}), \mathbf{f}_{2} (\mathbf{x}_2+\mathbf{o})\right\rangle},
\end{split}
\end{equation}
\end{footnotesize}

\noindent where $Corr (\mathbf{f}_{1},\mathbf{f}_{2})$ denotes the cost volume calculated via correlation. $\mathbf{o} \in [-n,n] \times [-n,n]$ denotes the offset of correlation operation with search region. In consideration of matrix multiplication in the attention mechanism of $f(X_i ,X_j )$ shown in Figure \ref{fig:twotypesmulti}, the different order of the two matrices in multiplication leads to great disparity of explanation with different displacements of correlation.

\noindent \emph{Discussions.} In Figure \ref{fig:twotypesmulti}(a), the expression is defined as $F_{2} (\mathbf{x}_2, c_2) (F_{1} (\mathbf{x}_1, c_1))^\top \in \mathbb{R}^{H_2 W_2 \times H_1 W_1 }$. If $F_1=F_2$, this operation strengthens the detail features of each position via aggregating information across channels from other positions, which would indicate the spatial attention integration at full resolution, and it is utilized to the basic non-local block \cite{wang2017non-local}. However, if $F_1 \ne F_2$, as the definition of cost volume, only the diagonal elements present the correlation with no displacement. On the contrary, the expression is defined as $(F_{1} (\mathbf{x}_1, c_1))^\top F_{2} (\mathbf{x}_2, c_2) \in \mathbb{R}^{C_1 \times C_2}$ in Figure \ref{fig:twotypesmulti}(b), which is a global correlation representation at full resolution among channels, and is essential to the naive correlation operation between feature pairs. For different matrix multiplication approaches, the attention maps catch dependencies with corresponding concepts in spatial features and temporal dynamics, which enhance representation for input feature extraction and correlation calculation, respectively.
%$\mathbf{f}_k$ is the flattened column vector of $F_k$.  $\sum\limits_{\mathbf{x}_1} \sum\limits_{\mathbf{x}_2} {(\mathbf{f}_{1} (\mathbf{x}_1, c_1))^\top \mathbf{f}_{2} (\mathbf{x}_2, c_2)}$so that the contextual representation

\begin{figure}[t]
 \centering\includegraphics[width=0.9\linewidth]{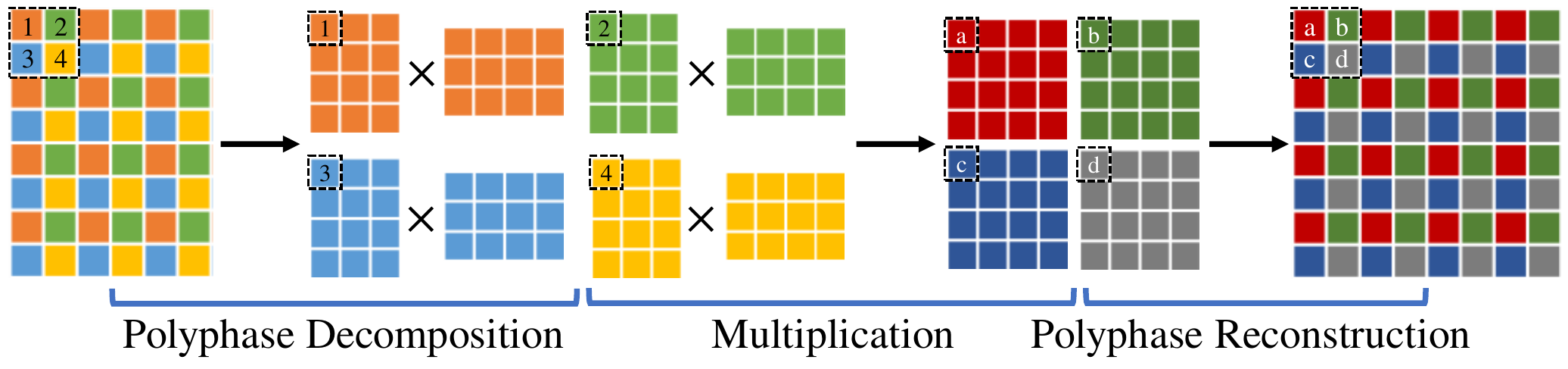}
 % \vspace{-2mm}
 \caption{The proposed simplified matrix multiplication with polyphase decomposition and reconstruction.}
 %\caption{The proposed simplified matrix multiplication (c) vs. (a) naive matrix multiplication, and (b) the downsampling-multiplication-upsampling scheme.}
 \label{fig:litemulti}
 % \vspace{-3mm}
\end{figure}

\noindent \emph{Lite matrix multiplication.} Considering the runtime of the flow prediction, the matrix multiplication in contextual attention block needs to be simplified with less computational complexity. In Figure \ref{fig:litemulti}, according to the neighbor similarity of images or frame pairs, we propose the polyphase decomposition and reconstruction scheme to simplify matrix multiplication opreation, which would obtain better approximation than the naive downsampling-upsampling scheme, and reduce the computation complexity compared to the direct multiplication. Denote the polyphase decomposition factor as $s$ ($s>1$). Given a reshaped feature  $\tilde F \in \mathbb{R}^{M \times N}$, the FLOPs of the entire multiplication is reduced from $O \left(NM^2 \right)$ to $O\left( \frac {NM^2}{s} \right)$. The comparison of different factors is presented in Section \ref{sec:ex}. %formulation
%\noindent \emph{Lite matrix multiplication.} Considered the runtime of the flow prediction, The matrix multiplication in contextual attention block need to be simplified with less computational complexity in Figure \ref{fig:litemulti}(c). According to the neighbor similarity of images or frame pairs, we propose the polyphase decomposition and reconstruction scheme for simplified matrix multiplication, which would obtain the better approximation than the downsampling-upsampling scheme in Figure \ref{fig:litemulti}(b), and reduce the computation complexity compared to the direct multiplication in Figure \ref{fig:litemulti}(a). Denote the polyphase decomposition factor as $s$ ($s>1$). Given a reshaped feature  $\tilde F \in \mathbb{R}^{M \times N}$, the FLOPs of the entire multiplication is reduced from $O \left(NM^2 \right)$ to $O\left( \frac {NM^2}{s} \right)$. The comparison of different factors is indicated in Section \ref{sec:ex}. %formulation
%Simplify non-local module.  Considered the scale factor of the , and t

\noindent\textbf{Contextual Attention Framework.} In general, the input of CNNs is not limited to the single feature through the single path, and the attention block needs to be adapted to more than one features, \eg two input features of the correlation operation. As shown in Figure \ref{fig:caf}(a), the components of the attention block can be abstracted as follows:

\emph{Attention aggregation.} To aggregate the attention integration feature to the intrinsic feature representation in each corresponding dimension, where the intrinsic representation often adopts basic operators like interpolation, convolution and transposed convolution.%, to fuse features for matching the effect of target operation.%, and generate the attention map adapted to the

\emph{Context transformation.} To transform the aggregated attention via the 1$\times$1 or 1D convolution, and obtain the contextual attention feature of all positions and channels. % via weighted averaging with weight j to obtain the global context features (global attention pooling in the simplified NL (SNL) block)%adopts a 1$\times$1 convolution Wk and softmax function to obtain the attention weights, and then performs the attention pooling to obtain the global context features

\emph{Target fusion.} To aggregate the output feature from target operation with the contextual attention, where the target operation is the main function to attain the objective from input features.
%\begin{itemize}
%  \item \emph{Attention aggregation.}  global attention pooling, which adopts a 1x1 convolution Wk and softmax function to obtain the attention weights, and then performs the attention pooling to obtain the global context features;
%  \item \emph{Context modeling.} feature transform via a 1x1 convolution Wv;
%  \item \emph{Target fusion.} feature aggregation, which employs addition to aggregate the global context features to the features of each position.
%\end{itemize}

Denote $X^{(k)}$ as the multiple input features. We regard this abstraction as a contextual attention framework defined as follows:
\begin{footnotesize}
\begin{equation}
\begin{split}
Z = \mathcal{G}\left( {T\left( {X^{(k)} } \right),\mathcal{F}\left( {A\left( {X^{(k)} } \right),\sum\limits_k {\omega _k X^{(k)} } } \right)} \right),
\end{split}
\end{equation}
\end{footnotesize}

\noindent where $\mathcal{F}(\cdot )$ and $\mathcal{G}(\cdot )$ are the fusion operations for attention aggregation and target fusion. $T(\cdot)$ and $A(\cdot)$ denote target operation and attention integration for the input features, $\omega$ is the factor of linear transformation. The non-local block or the other attention modules are the specific form of context attention block with the single input feature, \eg $A_{ij}\left( {X} \right) = {{f(X_i ,X_j )}}/{{\mathcal{N}(X)}}$, and $T\left( {X } \right)$ is the all-pass function in the non-local block.

%  ... The proposed contextual attention block
% To explore position and channel relation of features dependencies, DANet addresses a two-stream attention network for via self-attention mechanism.

\subsection{Pyramidal Spatial Context Module}
\begin{figure}[t]
 \centering\includegraphics[width=0.95\linewidth]{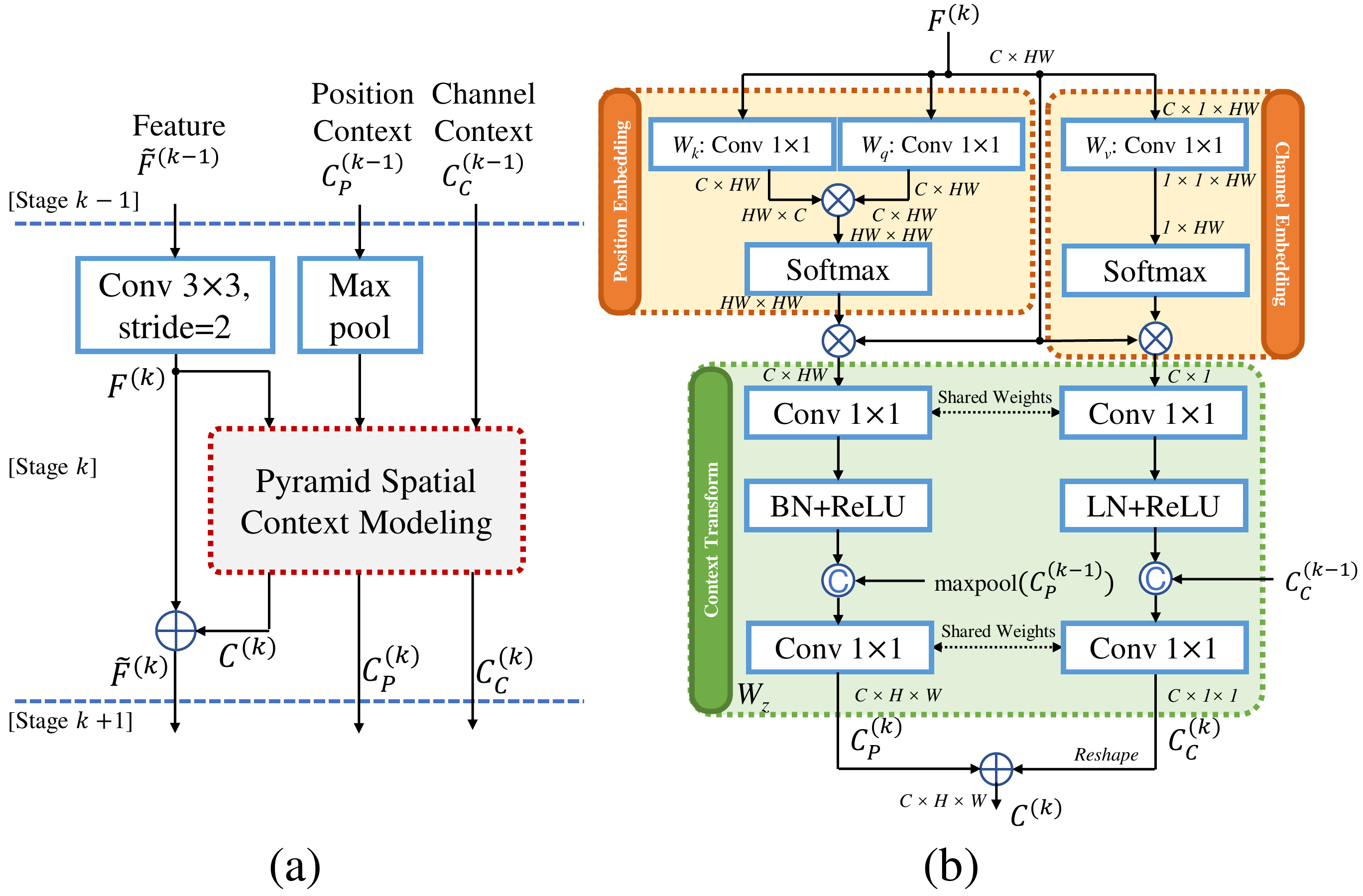}
 % \vspace{-2mm}
 \caption{The pyramidal spatial context (PSC) module. (a) The framework of PSC in the network; (b) The details of ``Pyramidal Spatial Context Modeling
'' in (a).}
 \label{fig:psc}
 % \vspace{-5mm}
\end{figure}

%Tight dual-att? Self-context modeling?
Inspired by the non-local network and global context network, we propose a pyramidal spatial context module with the tight dual-attention block to enhance the discriminative ability of feature representations in spatial position and channel dimensions. As shown in Figure \ref{fig:psc}, given a local feature $F^{(k)} \in \mathbb{R}^{C \times H \times W}$ at stage $k$, the calculation of the spatial context module is formulated as:

\begin{footnotesize}
\begin{equation}
\begin{split}
\tilde F^{(k)}  = F^{(k)}  + C_{P}^{(k)} + C_{C}^{(k)}, %W_z \sum\limits_j { \left(  C_{P(ij)} F_j + A_{C(ij)} F_j \right)},
\end{split}
\end{equation}
\end{footnotesize}
%\tilde F_i  = F_i  + W_z \sum\limits_j { \left(  A_{P(ij)} F_j + A_{C(ij)} F_j \right)},
\noindent where $C_{P}^{(k)}$ and $C_{C}^{(k)}$ are contextual attention at stage $k$ fused with that of stage $k-1$, which is to aggregate context from different granularity:

\begin{footnotesize}
\begin{equation}
\begin{split}
C_{P}^{(k)} &= W_z^{(k)} \left[ \sum\limits_j {  A_{P,ij}^{(k)} F_j^{(k)}}, C_{P}^{(k-1)} \Downarrow \right], \\
C_{C}^{(k)} &= W_z^{(k)} \left[ \sum\limits_j {  A_{C,ij}^{(k)} F_j^{(k)}}, C_{C}^{(k-1)} \right], %W_z \sum\limits_j { \left(  C_{P(ij)} F_j + A_{C(ij)} F_j \right)},
\end{split}
\end{equation}
\end{footnotesize}

\noindent where ``$\Downarrow$'' denotes max-pooling, and ``$\left[ \cdot \right]$'' denotes the concatenation operator. $A_{P,ij}$ and $A_{C,ij}$ are attention integrations in position and channel, defined as follows to learn the spatial and channel interdependencies:

\begin{footnotesize}
 \begin{equation}
\begin{split}
%\begin{cases}
  A_{P,ij} \left( {F} \right) &= \frac{{\exp((W_q F_i)^{\top}(W_k F_j ))}}{{\sum\limits_m {\exp((W_q F_i)^{\top}(W_k F_m ))}}} \in \mathbb{R}^{HW \times HW}, \\
  A_{C,ij} \left( {F} \right) &= \frac{{\exp(W_k F_j )}}{{\sum\limits_m {\exp(W_k F_m )}}} \in \mathbb{R}^{HW \times 1}.
%\end{cases}
\end{split}
\end{equation}
\end{footnotesize}

%and
%\noindent self-context learning

%we first feed it into a convolution layers to generate
%two new feature maps B and C, respectively, where
%fB;Cg 2 RCHW. Then we reshape them to RCN,
%where N = H  W is the number of pixels. After that we
%perform a matrix multiplication between the transpose of C
%and B, and apply a softmax layer to calculate the spatial
%attention map S 2 RNN:

%A position attention module is proposed to learn the spatial interdependencies of features and a channel attention module is designed to model channel interdependencies. It significantly improves the results by modeling rich contextual dependencies over local features.

\subsection{Temporal Context Correlation Module}
\begin{figure}[t]
 \centering\includegraphics[width=\linewidth]{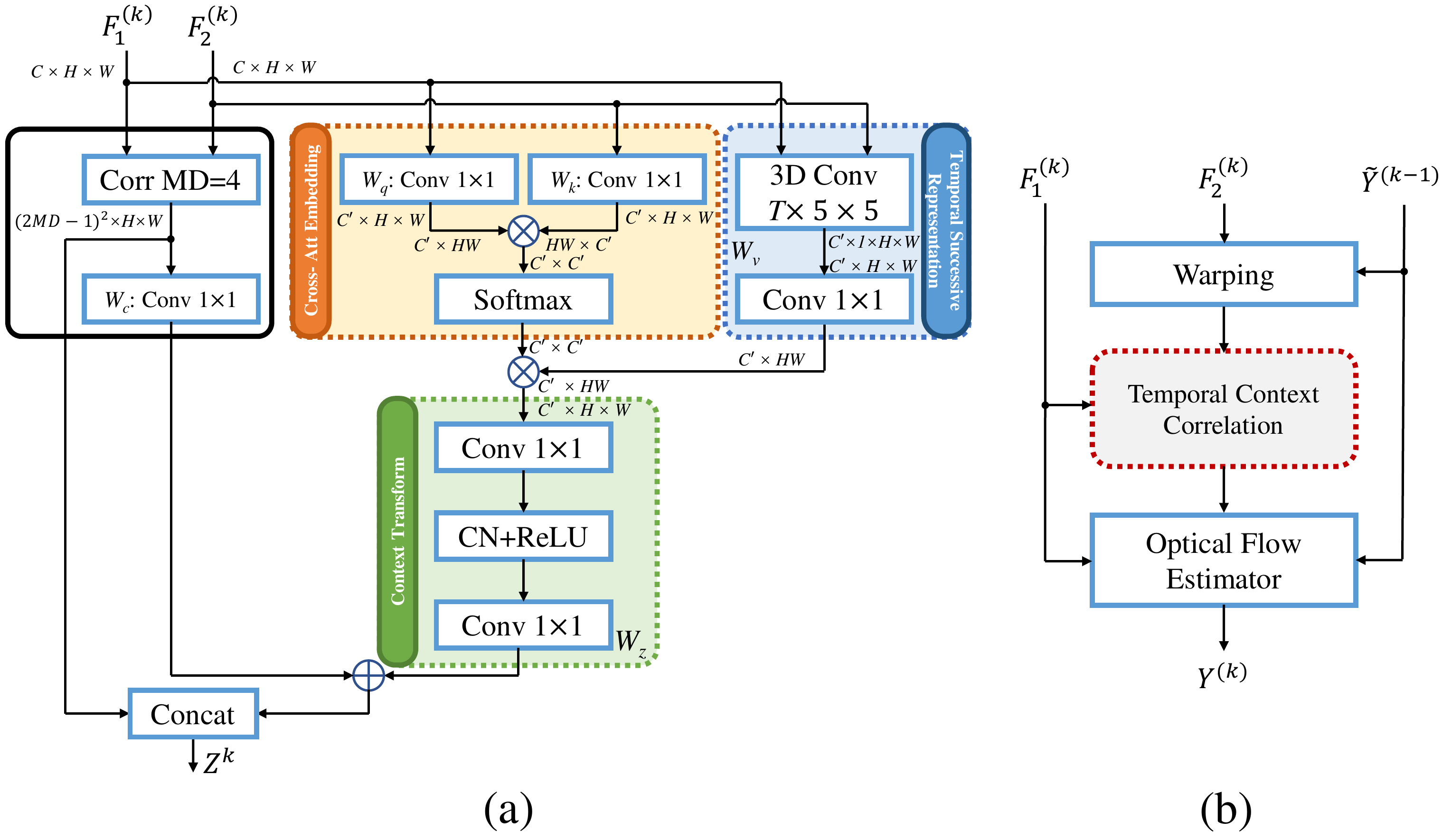}
 %% \vspace{-2mm}
 \caption{The temporal context correlation (TCC) module. (a) The details of TCC module; (b) The Contextual PWC Module utilized with TCC in (a). ``MD'' is the max displacement of correlation. The temporal successive representation utilizes 3D convolution with kernel size $T\times 5\times 5$, and $T$ is the frame number of input, \ie 2.}
 \label{fig:tcc}
 % \vspace{-6mm}
\end{figure}

%Cross Context fusion?
As the spatial context module learns query-independent context relationships at the feature extraction stage, the temporal context module is adopted to model the relationships of correlation calculation. As the analysis on matrix multiplication, full-resolution correlation is utilized to describe the global context of correlation operation. As shown in Figure \ref{fig:tcc}(a), given the local feature pairs $F_1, F_2 \in \mathbb{R}^{C \times H \times W}$ from feature extraction, the contextual correlation is formulated as:

\begin{footnotesize}
\begin{equation}
\begin{split}
Z_i  = W_c Corr_i (F_1,F_2)  + W_z \sum\limits_j { \left( A_{T,ij} \cdot W_v (F_1,F_2) \right)},
\end{split}
\end{equation}
\end{footnotesize}

\noindent where $A_{T,ij}$ is the temporal attention integration with the ``cross-attention'' mechanism, which is defined as follows:

\begin{footnotesize}
 \begin{equation}
\begin{split}
%\begin{cases}
  A_{T,ij} \left( {F_1,F_2} \right) &= \frac{{\exp((W_q F_{1,i})^{\top}(W_k F_{2,j} ))}}{{\sum\limits_m {\exp((W_q F_{1,i})^{\top}(W_k F_{2,m} ))}}} \in \mathbb{R}^{C \times C}.
%\end{cases}
\end{split}
\end{equation}
\end{footnotesize}

Notice that the linear transformation of $W_v (F_1,F_2)$ is modeled by a 3D convolution and a 1$\times$1 convolution, which aims to explore the temporal information across time dimension. Since the max displacement of correlation is selected to 4, the kernel of 3D convolution needs to cover all frames in the temporal dimension, and the height and width are greater than or equal to the max displacement, \ie 5 in the proposed module.

The TCC module is a flexible correlation operator and it can be utilized to PWC module in PWC-Net \cite{Sun_2018_CVPR} as ``Contextual PWC'' module, to learn long-dependencies between the reference feature and the warped feature.

\subsection{Recurrent Residual Contextual Upsampling}
\begin{figure}[t]
 \centering\includegraphics[width=0.9\linewidth]{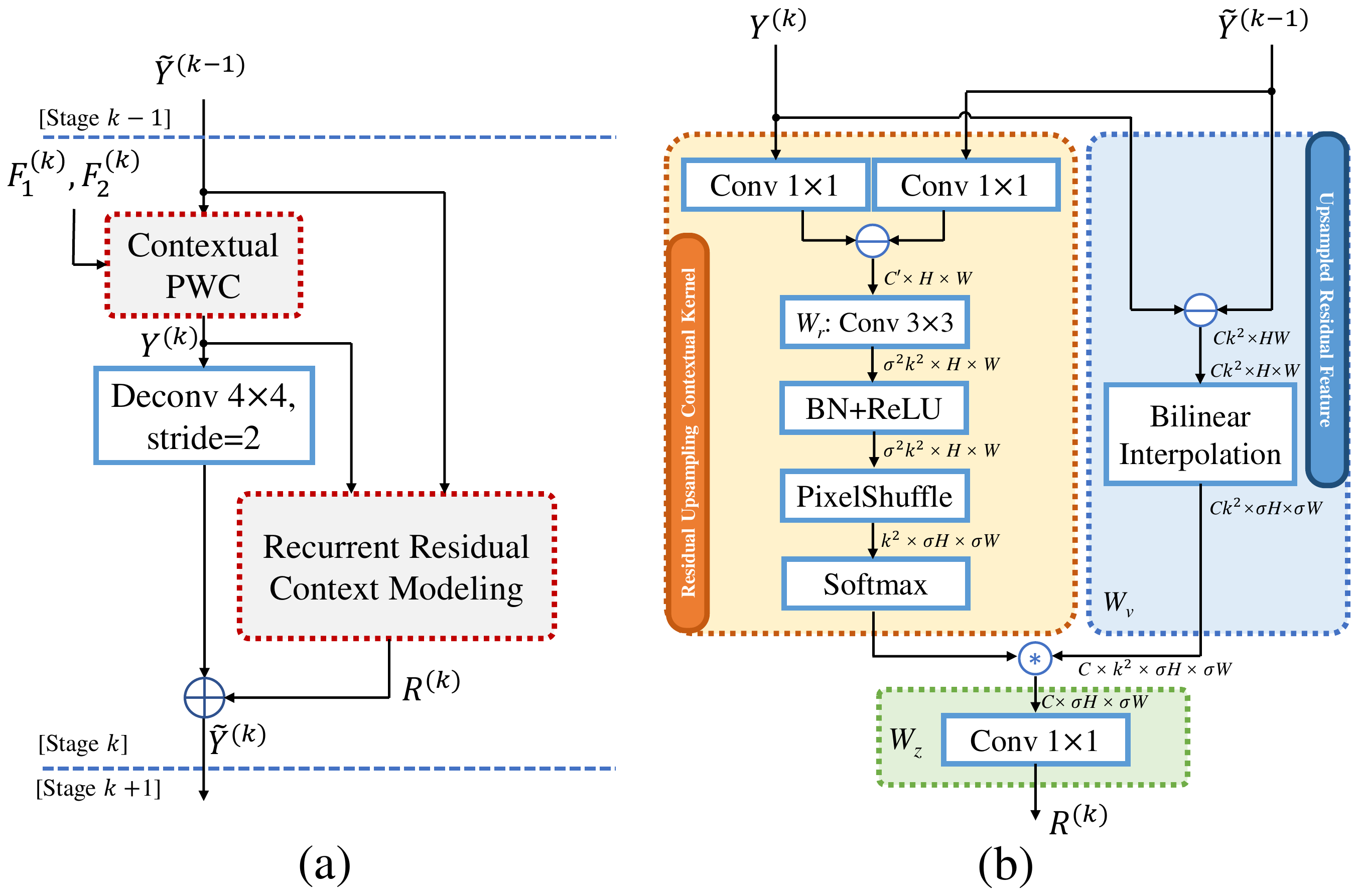}
 % \vspace{-2mm}
 \caption{The recurrent residual contextual upsampling (RRCU) module. (a) The framework of RRCU in the network, with the Contextual PWC module in Figure \ref{fig:tcc}; (b) The details of ``Recurrent Residual Context Modeling'' in (a).}
 \label{fig:upsampling}
 % \vspace{-6mm}
\end{figure}

Different from the spatial and temporal context representation modeling, the reconstruction context learning is a detail-aware operation to learn high-frequency features and preserve edges within a large receptive field. In view of the multi-stage structure of reconstruction, we propose an efficient recurrent module for upsampling, which leverages the underlying content information between the current stage and the previous stage.

The predicted optical flow $Y^{(k)}$ at stage $k$ and the upsampled optical flow $\tilde Y^{(k-1)}$ at stage $k-1$ are encoded by 1$\times$1 convolution with shared weights, and the residual with smaller size $R^{(k)}$ is calculated from the encoder at first. Denote the residual between $Y^{(k)}$ and $\tilde Y^{(k-1)}$ as $R^{(k)} = Y_i^{(k)}-\tilde Y_i^{(k-1)}$, and then the context modeling is utilized for $R^{(k)}$ to explore the up-sampling attention kernels $A_U$ for each corresponding source position, and $A_U$ is fused back to the bilinear interpolated $R^{(k)}$. Finally, the fined residual feature $\tilde R^{(k)}$ is resembled to $Y^{(k)}$ to obtain the refined upsampled flow $\tilde Y^{(k)}$ with rich details. The architecture is illustrated in Figure \ref{fig:upsampling}, and the formulation is expressed as follows:

\begin{footnotesize}
\begin{equation}
\begin{split}
\tilde Y_i^{(k)}  = deconv (Y_i^{(k)})  + W_z \sum\limits_i { \left( A_{U,i} * W_v R_i^{(k)} \right)},
\end{split}
\end{equation}
\end{footnotesize}

\noindent where ``*'' denotes the position-wise convolution operator, and here $W_v$ is a bilinear interpolation operator for $R^{(k)}$. $A_{U,ij}$ denotes the adaptive attention kernels to model the detail context defined as follows:

%use an adaptive and optimized reassembly kernel in different locations learn the spatial and channel interdependencies with the ``cross-attention'' mechanism, which is defined as follows:
\begin{footnotesize}
 \begin{equation}
\begin{split}
%\begin{cases}
  A_{U,i} \left( R \right) &= \frac{{\exp(ps(W_r R_i))}}{{\sum\limits_m {\exp(ps(W_r R_m))}}} \in \mathbb{R}^{\sigma ^2 \times H \times W},
%\end{cases}
\end{split}
\end{equation}
\end{footnotesize}

\noindent where $ps$ denotes the ``Pixel Shuffle \cite{shi2016real-time}'' operator for sub-pixel convolution, to reconstruct the sub-pixel information and preserve edges and textures. $\sigma$ is the upsampling factor, and here $\sigma=2$.

\subsection{Overall Architecture}

Given the proposed contextual attention modules, we now describe the overall architecture of the proposed STC-Flow. The input is the frame pairs $I_1$ and $I_2$ with size $3 \times H \times W$, and the goal of STC-Flow is to obtain the optical flow map $Y$ with size $2\times H \times W$. The contextual representations are modeled via three key components --- pyramidal spatial context (PSC) module, temporal context correlation (TCC) module, and recurrent residual contextual upsampling (RRCU) module, to leverage long-range dependencies relationship in feature extraction, correlation and flow reconstruction, respectively. The entire network is trained jointly, shown in Figure \ref{fig:arch}. % We propose an end-to-end spatio-temporal context-aware architecture for optical flow estimation

Since PWC-Net \cite{Sun_2018_CVPR} and LiteFlowNet \cite{hui2018liteflownet} provide superior performance with lightweight architectures, we take a simplified version of PWC-Net, with layer reduction in feature extraction and reconstruction, as the baseline of our STC-Flow. For successive of image/frame pairs, the backbone network with PSC outputs pyramidal feature maps for each image. With the feature maps of each stage converted to cost volumes via correlation operation, the cost volumes are decoded and reconstructed to predict optical flow, assisted by TCC. With the guidance of backbone features and warping alignments, the predicted flow field goes through the RRCU module and the fined flow is obtained.

\section{Experiments}
\label{sec:ex}

In this section, we introduce the implementation details, and evaluate our method on public optical flow benchmarks, including MPI Sintel \cite{Butler2012A}, KITTI 2012 \cite{Geiger2012Are} and KITTI 2015 \cite{Menze2015Object}, and compare it with state-of-the-art methods.

\subsection{Implementation and training details}

%\noindent\textbf{Network architecture. }

%% \vspace{1ex}

%\noindent\textbf{Training procedures. }
We take a simplified version of PWC-Net, with the same number of stages and layer reduction in feature extraction and reconstruction. PSC and RRCU modules are utilized at stage 3, 4 and 5 for feature extraction and reconstruction respectively. TCC Module is applied at stage 3, 4, 5 and 6 for correlation of feature pairs or warped features. The training loss weights among stages are 0.32, 0.08, 0.02, 0.01, 0.005. We first train the models with the FlyingChairs dataset \cite{dosovitskiy2015flownet} using L2 loss and the $S_{long}$ learning rate schedule, with random flipping and cropping of size 448 $\times$ 384 introduced by \cite{Ilg_2017_CVPR}. Secondly, we fine-tune the models on the FlyingThings3D dataset \cite{Mayer2016A} using the $S_{fine}$ schedule with cropping size of 768 $\times$ 384. Finally, the model is fine-tuned on Sintel and KITTI datasets using the general Charbonnier function $\rho(x)=\left( x^2 + \epsilon ^2 \right)^q$ ($q<1$) as the robust training loss. We use both the \emph{clean} and \emph{final} pass of the training data throughout the Sintel fine-tuning process, with cropping size of 768 $\times$ 384; and we use the mixed data of KITTI 2012 and 2015 training for KITTI fine-tuning process, with cropping size of 896 $\times$ 320. % For the ground truth of KITTI dataset is semi-dense, larger patches could capture the large motion. %The trade-off weight $\gamma$ of loss fuction $\mathcal{L}$ is set to 0.0005. We scale the ground truth flow by 20, the same as \cite{bibid}. For each of the six levels, we downsample the ground truth by a factor of $2^n$, where $n$ denotes the index of the stages., random flipping, random channel swapping  In terms of the network architecture, we train our network by the following steps.

\begin{table}[t]
%\centering
	\caption{Ablation study of our component choices of the network. Average end-point error (AEE) and percentage of erroneous pixels (Fl-all) Results of our STC-Flow with different components of PSC, TCC and RRCU on Sintel training \emph{Clean} and \emph{Final} passes, and KITTI \emph{2012}/\emph{2015}.}

    % \vspace{-1.5mm}
\begin{footnotesize}
	\begin{subtable}[h]{\linewidth}
		%\centering
		\caption{\textbf{Pyramidal Spatial Context Module} improves quantity results significantly. ``w. PSC$_{3-5}$'' means ``using PSC in stage 3, 4 and 5''}
        % \vspace{-1.5mm}	
		\label{table:SCM}
        \begin{tabular}{p{1.7cm}p{0.7cm}<{\centering}p{0.7cm}<{\centering}p{1.4cm}<{\centering}p{0.6cm}<{\centering}p{0.8cm}<{\centering}}
			\toprule
			~& \multicolumn{2}{c}{Sintel} & {KITTI 2012} & \multicolumn{2}{c}{KITTI 2015}\\
			~ & Clean & Final & AEE  & AEE & Fl-all \\
			\hline
			baseline & 2.924 & 4.088 & 4.621 & 11.743 & 36.53\%\\
			w. PSC$_{3}$ & 2.802 & 3.891 & 4.565 & 11.031 & 35.37\% \\
			w. PSC$_{3-4}$ & 2.747 & 3.873 & 4.545 & 10.677 & 34.84\% \\
			w. PSC$_{3-5}$ & 2.741 & 3.864 & 4.494 & 10.332 & 34.45\% \\
			w. 2D-NL$_{3-5}$ & 2.785 & 3.968 & 4.523 & 10.482 & 34.76\%  \\
            Full model &  2.412 &  3.601 &4.196 &  10.181 & 32.23\%  \\
			\bottomrule
		\end{tabular}
	\end{subtable}
\end{footnotesize}

    % \vspace{2.2mm}

\begin{footnotesize}
	\begin{subtable}[h]{\linewidth}
		\centering

		\caption{\textbf{Temporal Context Correlation Module} is critical and outperforms single correlation module.}\label{table:TCCM}
        % \vspace{-1.5mm}			
        \begin{tabular}{p{1.7cm}p{0.7cm}<{\centering}p{0.7cm}<{\centering}p{1.4cm}<{\centering}p{0.6cm}<{\centering}p{0.8cm}<{\centering}}
        %\begin{tabular}{p{1.4cm}p{0.7cm}<{\centering}p{0.7cm}<{\centering}p{1.6cm}<{\centering}p{0.7cm}<{\centering}p{0.8cm}<{\centering}}
			\toprule
			~& \multicolumn{2}{c}{Sintel} & {KITTI 2012} & \multicolumn{2}{c}{KITTI 2015}\\
			~ & Clean & Final & AEE  & AEE & Fl-all \\
			\hline
			baseline & 2.924 & 4.088 & 4.621 & 11.743 & 36.53\%\\
			w. TCC$_{6}$ & 2.787 & 3.863 & 4.523 & 10.712 & 35.59\%\\
			w. TCC$_{3-6}$ & 2.641 & 3.780 & 4.389 & 10.313 & 34.58\%\\
			w. 2D-NL$_{3-6}$ & 2.764 & 3.869 & 4.498 & 10.564 & 35.25\%\\
			w. 3D-NL$_{3-6}$ & 2.635 & 3.745 & 4.393 & 10.324 & 34.63\%\\
            Full model &  2.412 &  3.601 &4.196 &  10.181 & 32.23\%  \\
			\bottomrule
		\end{tabular}
	\end{subtable}
\end{footnotesize}

    % \vspace{2.2mm}

\begin{footnotesize}
	\begin{subtable}[h]{\linewidth}
		\centering
		\caption{\textbf{Recurrent Residual Context Upsampling} has better performance.}	
        % \vspace{-1.5mm}	
        \begin{tabular}{p{1.7cm}p{0.7cm}<{\centering}p{0.7cm}<{\centering}p{1.4cm}<{\centering}p{0.6cm}<{\centering}p{0.8cm}<{\centering}}
        %\begin{tabular}{p{1.4cm}p{0.7cm}<{\centering}p{0.7cm}<{\centering}p{1.6cm}<{\centering}p{0.7cm}<{\centering}p{0.8cm}<{\centering}}
			\toprule
			~& \multicolumn{2}{c}{Sintel} & {KITTI 2012} & \multicolumn{2}{c}{KITTI 2015}\\
			~ & Clean & Final & AEE  & AEE & Fl-all \\
			\hline
			baseline & 2.924 & 4.088 & 4.621 & 11.743 & 36.53\%\\
			w. RRCU & 2.696 & 3.794 & 4.432 & 10.332 & 34.65\% \\
			TCC+RRCU & 2.567 & 3.722 & 4.368 & 10.295 & 33.89\% \\
            Full model &  2.412 &  3.601 &4.196 &  10.181 & 32.23\%  \\
			\bottomrule
		\end{tabular}
		\label{table:RRCU}

	\end{subtable}
\end{footnotesize}
	\label{table:ablationstudy}
% \vspace{-5mm}	
\end{table}

\subsection{Ablation Study}

\begin{figure*}[t]
 \centering\includegraphics[width=\textwidth]{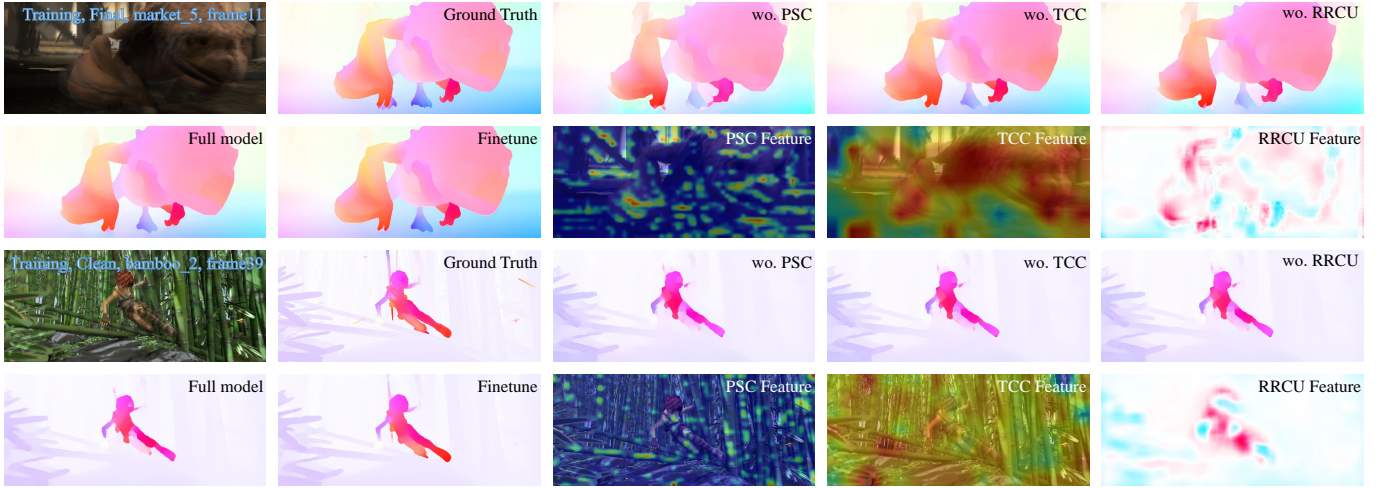}
 % \vspace{-6mm}
 \caption{Results of ablation study on Sintel training \emph{Clean} and \emph{Final} passes. We also indicate the learned features on corresponding modules --- PSC and RRCU in stage 4 and TCC in stage 6. (Zoom in for details.)}
 \label{fig:ablation}
 % \vspace{-0.7mm}
\end{figure*}

To demonstrate the effectiveness of individual contextual attention module in our network, as shown in Table \ref{table:ablationstudy} and Figure \ref{fig:ablation}, we conduct a rigorous ablation study of PSC, TCC, and RRCU, respectively. We observe that these modules could capture clear semantic information with long-range dependencies. The baseline is trained on FlyingChairs and finetuned on FlyingThings3D. We also discuss the efficacy of Lite matrix multiplier in Table \ref{table:lite}.

%% \vspace{1ex}
\noindent\textbf{Pyramidal spatial context module. }STC-Flow utilizes PSC Module in level 3, 4 and 5. Table \ref{table:ablationstudy}(a) demonstrates that using PSC Module can improve the performance on both the Sintel and KITTI datasets, since this module enhances the ability of discriminating feature texture in feature extraction stage, and PSC at stage 3 is more beneficial, for the low-level discriminative details matter.

%% \vspace{1ex}
\noindent\textbf{Temporal context correlation module. }TCC Module describes the relationship of correlation with the spatial and temporal context. In Table \ref{table:ablationstudy}(b), we compare the performance of our network using TCC Module with naive correlation operator, and also compare with 2D non-local block for concatenated feature and 3D non-local block for feature pairs. It demonstrates that fusion of correlation with spatial and temporal context is better than single correlation. Notice that 3D non-local blocks perform better in Sintel, however, with heavy computational complexity. TCC can achieve the comparable performance with fewer FLOPs.

%% \vspace{1ex}
\noindent\textbf{Recurrent residual contextual upsampling. }We utilize the RRCU Module to learn high-frequency context features and preserve edges. In Table \ref{table:ablationstudy}(c), we compare the quantity of our method using RRCU with single transpose convolution, which demonstrates that reconstruction context learning could preserve details and improve performance.

%% \vspace{1ex}
\noindent\textbf{Lite matrix multiplication. }Lite matrix multiplication is an efficient scheme to reduce the computational complexity. We compare the performance of this scheme with different polyphase decomposition factor $s$ on Sintel training. As shown in Table \ref{table:lite}, lite matrix multiplication has a margin influence on AEE, but increases the frame rate conspicuously. Considering the tradeoff between accuracy and time consumption, we select $s=2$ for the full model.%, in terms of average end-point error (AEE) and frame rates

%%% Non-local
%% 1,2,3,4 time epe difference
\begin{table}[t]
\caption{Detailed results of lite matrix multiplication with different polyphase decomposition factor $s$ on Sintel training \emph{clean} and \emph{final} pass dataset on AEE and frame rate, and structural similarity index (SSIM) of context features in stage 4 between lite multiplication and naive multiplication. (Inference on Intel Core i5 CPU and NVIDIA GEFORCE 1080 Ti GPU for the frame rate.)}
% \vspace{-5mm}
	\begin{center}
		\label{table:lite}
		\fontsize{8pt}{10pt}\selectfont\centering
		\begin{tabular}{p{0.2cm}p{2.4cm}<{\centering}p{2.4cm}<{\centering}p{1.6cm}<{\centering}}
        %\begin{tabular}{lcccc}
			\toprule
			$s$& AEE/SSIM (\emph{Clean}) & AEE/SSIM (\emph{Final}) & Runtime (fps)\\
			%~ & ~ & ~ & ~ & Final & Clean\\
			\hline
			1 & 2.407/--- & 3.588/--- & 20   \\
			2 & 2.412/0.9765 &  3.601/0.9982 & 22 \\
			4 & 2.515/0.9061 & 3.856/0.8990 & 25 \\
			\bottomrule
		\end{tabular}
	\end{center}
% \vspace{-5mm}
\end{table}

\begin{figure*}[t]
 \centering\includegraphics[width=\textwidth]{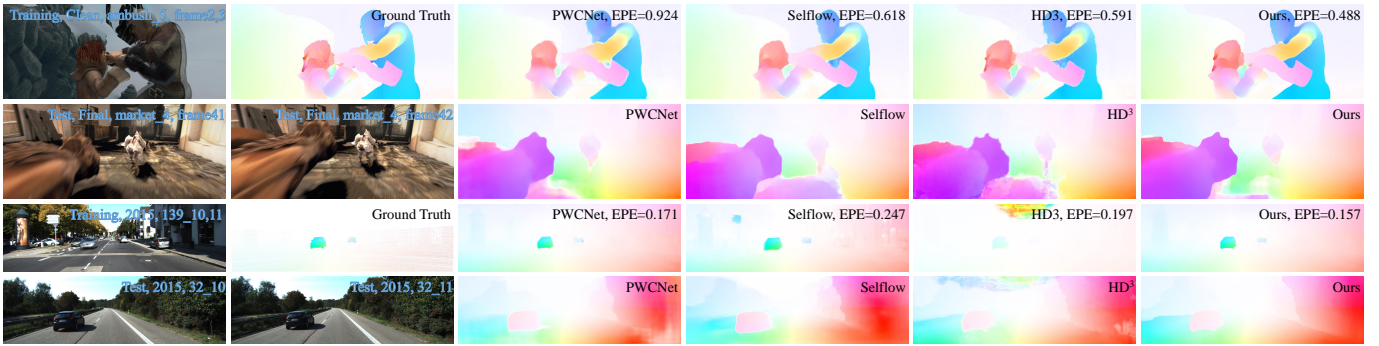}
 % \vspace{-6mm}
 \caption{Examples of predicted optical flow from different methods on Sintel and KITTI datasets. Our method achieves the better performance and preserves the details with fewer artifacts.  (Zoom in for details.)}
 \label{fig:compare}
 % \vspace{-0.7mm}
\end{figure*}

\begin{table*}[t]
\caption{AEE and Fl-all of different methods on Sintel and KITTI datasets. The ``-ft'' suffix denotes the fine-tuned networks using the target dataset. The values in parentheses are the results of the networks on the data they were trained on, and hence are not directly comparable to the others.} %
% \vspace{-5.5mm}
\fontsize{9pt}{10pt}\selectfont\centering
   \begin{center}
	\label{mainresults}
	%\begin{threeparttable}
	%\tiny %%scriptsize
	\fontsize{8pt}{9.5pt}\selectfont\centering
	\begin{tabular}{lp{1.1cm}<{\centering}p{1.1cm}<{\centering}p{1.1cm}<{\centering}p{1.1cm}<{\centering}p{1.1cm}<{\centering}p{1.1cm}<{\centering}p{1.1cm}<{\centering}p{1.4cm}<{\centering}p{1.2cm}<{\centering}}%{ccccccccc}	%{cp{0.4cm}cp{0.4cm}cp{0.4cm}cp{0.4cm}c|}%{l|cc|c}	
	\toprule  %添加表格头部粗线
	\multirow{2}*{Method}& \multicolumn{2}{c}{Sintel \emph{Clean}} & \multicolumn{2}{c}{Sintel \emph{Final}}  & \multicolumn{2}{c}{KITTI 2012} & \multicolumn{3}{c}{KITTI 2015}\\
	%Method & \multicolumn{2}{c}{Sintel clean} & \multicolumn{2}{c}{Sintel final}  & \multicolumn{2}{c}{KITTI 2012} & \multicolumn{2}{c}{KITTI 2015}\\
	~ & train & test & train & test & train & test & train &train(Fl-all)& test(Fl-all)\\
	\hline
	%LDOF~\cite{brox2010large} & 4.64 & 7.56 & 5.96 & 9.12 & 10.94 & 12.4 & 18.19 & 38.11\% & --- \\
	DeepFlow~\cite{Weinzaepfel2013DeepFlow} & 2.66 & 5.38 & 3.57 & 7.21 & 4.48 & 5.8 & 10.63 & 26.52\% & 29.18\%  \\
	%PCA-Layers~\cite{wulff2015efficient} & 3.22 & 5.73 & 4.52 & 7.89 & 5.99 & 5.2 & 12.74 & 27.26\% & --- \\
	EpicFlow~\cite{revaud2015epicflow} & 2.27 & 4.12 & 3.56 & 6.29 & 3.09 & 3.8 & 9.27 & 27.18\% & 27.10\%  \\
	FlowFields~\cite{bailer2015flow} & 1.86 & 3.75 & 3.06 & 5.81 & 3.33 & 3.5 & 8.33 & 24.43\% & ---\\
	%Full Flow~\cite{Chen2016Full} & --- & \textbf{2.71} & 3.60 & 5.90 &&&&&\\
	%\hline
	%Deep DiscreteFlow~\cite{guney2016deep} & --- & 3.86 & --- & 5.73 & --- & 3.4 & --- & --- & 21.17\% \\
	%Patch Matching~\cite{bailer2017cnn} & --- & 3.78 & --- & 5.36 & --- & 3.0 & --- & --- & 19.44\%\\
	%DC Flow~\cite{Xu_2017_CVPR} & --- & --- & --- & \textbf{5.12} & --- & --- & --- & --- & 14.86\%\\
	%\hline
	FlowNetS~\cite{dosovitskiy2015flownet} & 4.50 & 7.42 & 5.45 & 8.43 & 8.26 & --- & --- & --- & --- \\
	FlowNetS-ft~\cite{dosovitskiy2015flownet} & (3.66) & 6.96 & (4.44) & 7.76 & 7.52 & 9.1 & --- & --- & --- \\
	FlowNetC~\cite{dosovitskiy2015flownet} & 4.31 & 7.28 & 5.87 & 8.81 & 9.35 & --- & --- & --- & --- \\
	FlowNetC-ft~\cite{dosovitskiy2015flownet} & (3.78) & 6.85 & (5.28) & 8.51 & 8.79 & --- & --- & --- & ---\\
	FlowNet2~\cite{Ilg_2017_CVPR} & 2.02 & 3.96 & 3.54 & 6.02 & 4.01 & --- & 10.08 & 29.99\% & --- \\
	FlowNet2-ft~\cite{Ilg_2017_CVPR} & (1.45) & 4.16 & (2.19) & 5.74 & 3.52 & --- & 9.94 & 28.02\% & ---\\
	%\hline
	SPyNet~\cite{ranjan2017optical} & 4.12 & 6.69 & 5.57 & 8.43 & 9.12 & --- & --- & --- & ---\\
	SPyNet-ft~\cite{ranjan2017optical} & (3.17) & 6.64 & (4.32) & 8.36 & 3.36 & 4.1 & --- & --- & 35.07\%\\

	LiteFlowNet~\cite{hui2018liteflownet} & 2.48 & --- & 4.04 & --- & 4.00 & --- & 10.39 & 28.50\% & ---\\
	LiteFlowNet-ft~\cite{hui2018liteflownet} & (1.35) & 4.54 & (1.78) & 5.38 & (1.05) & 1.6 & (1.62) & (5.58\%) & (9.38\%) \\
	%LiteFlowNet2~\cite{hui2018liteflownet} & 2.24 & --- & 3.78 & --- & 3.42 & --- & 8.97 & 25.88\% & ---\\
	%LiteFlowNet2-ft~\cite{} & (1.30) & 3.45 & (1.62) & 4.90 & (1.00) & 1.4 & (1.47) & 4.80\% & 7.74\%\\
	PWC-Net~\cite{Sun_2018_CVPR} & 2.55 & --- & 3.93 & --- & 4.14 & --- & 10.35 & 33.67\% & --- \\
	PWC-Net-ft~\cite{Sun_2018_CVPR} & (2.02) & 4.39 & (2.08) & 5.04 & (1.45) & 1.7 & (2.16)  & (9.80\%) & 9.60\%\\
	%PWC-Net-ft-arXiv~\cite{} & (1.71) & 3.45 & (2.34) & 4.60 & (1.08) & 1.5 & 6.82\%  & (1.45) & 7.90\%\\
	%ProFlow-ft~\cite{} & (1.78) & 2.82 & --- & 5.02 & (1.89) & 2.1 & 7.88\% & (5.22) & 15.04\%\\
	%ContinualFlow-ft~\cite{} & --- & 3.34 & --- & 4.52 & --- & --- & --- & --- & 10.03\%\\
	SelFlow-ft~\cite{liu2019selflow} & (1.68) & 3.74 & (1.77) & \textbf{4.26} & (0.76) & 1.5 & (1.18) & --- & 8.42\%\\
	IRR-PWC-ft~\cite{hur2019iterative} & (1.92) & 3.84 & (2.51) & 4.58 & --- & --- & (1.63) & (5.32\%) & 7.65\%\\
	HD3-ft~\cite{yin2018hierarchical} & (1.70) & 4.79 & (1.17) & 4.67 & (0.81) & \textbf{1.4} & (1.31) & (4.10\%) & \textbf{6.55\%}\\
	\hline
	STC-Flow (ours) & 2.41 & --- & 3.60 & --- & 4.20 & --- & 10.18 & 32.23\% & --- \\
	STC-Flow-ft (Ours) & (1.36) & \textbf{3.52} & (1.73) & 4.87 & (0.98) & 1.5 & (1.46) & (5.43\%) & 7.99\%\\
	\bottomrule %添加表格底部粗线
\end{tabular}

\end{center}

\end{table*}

\subsection{Comparison with State-of-the-art Methods}

As shown in Table \ref{mainresults}, we achieve the comparable quantity results in Sintel and KITTI datasets compared with state-of-the-art methods. Some samples of visualization results are shown in Figure \ref{fig:compare}. STC-Flow performs better on AEE among the methods on the Sintel \emph{Clean} pass. We can see that the finer details are well preserved via context modeling of spatial and temporal long-range relationships, with fewer artifacts and lower end-point error. In addition, our method is based on only two frames without additional information (like occlusion maps for IRR \cite{hur2019iterative} and additional datasets) used, but it outperforms state-of-the-art multi-frames methods, \eg SelFlow\cite{liu2019selflow}. In addition, STC-Flow is lightweight with far fewer parameters, \ie 9M instead of 110M of FlowNet2 \cite{Ilg_2017_CVPR} and 40M of HD$^3$ \cite{yin2018hierarchical}. We believe that our flexible scheme is helpful to achieve better performance for other baseline networks, including multi-frame based methods.%However, the motion of tiny or slender objects is challenging, where edges and patches are not preserved effectively, and we will perform further study on the more challenging cases.

%% \vspace{1ex}

%\noindent\textbf{KITTI.} The size of the sequences is not same in KITTI dataset. During inference and test period, we upsample the input image pairs to a multiple of 64 and output a flow in the same size. Then we resize the output flow to the original size of the image pairs. STC-Flow achieves the state-of-art results on KITTI dataset, with Fl-all=... on KITTI 2012 and Fl-all=... on KITTI 2015 and outperforms all published two-frame optical flow methods.

\section{Conclusion}

To explore the motion context information for accurate optical flow estimation, we propose a spatio-temporal context-aware network, STC-Flow, for optical flow estimation. We propose three context modules for feature extraction, correlation, and optical flow reconstruction stages, \ie \emph{pyramidal spatial context  (PSC) module}, \emph{temporal context correlation  (TCC) module}, and \emph{recurrent residual contextual upsampling (RRCU) module}, respectively. We have validated the effectiveness of each component. Our proposed scheme achieves the state-of-the-art performance without multi-frame or additional information used.
% References and End of Paper

% or


%\appendices
%% you can choose not to have a title for an appendix
%% if you want by leaving the argument blank
%\section{}

%\section*{Acknowledgment}

\ifCLASSOPTIONcaptionsoff
  \newpage
\fi



\bibliographystyle{IEEEtran}
% �ο�����
\bibliography{reference}
\end{document}